\documentclass{article}

\usepackage[final,nonatbib]{nips_2018}

\usepackage{graphicx}
\usepackage{color,xcolor}

\usepackage{adjustbox}
\usepackage{array}
\usepackage{booktabs}
\usepackage{lipsum}
\usepackage{colortbl}
\usepackage{float,wrapfig}
\usepackage{hhline}
\usepackage{multirow}

\usepackage{amsmath,amsfonts,amsthm,amssymb,mathtools}
\usepackage{bm}
\usepackage{nicefrac}
\usepackage{microtype}

\usepackage{changepage}
\usepackage{extramarks}
\usepackage{fancyhdr}
\usepackage{lastpage}
\usepackage{setspace}
\usepackage{soul}
\usepackage{xspace}

\usepackage[pagebackref=true,breaklinks=true,colorlinks,citecolor=gray]{hyperref}
\usepackage{url}

\usepackage{algorithm, algorithmic}
\usepackage{enumerate}
\usepackage{todonotes} 

\usepackage[compact]{titlesec}

\usepackage[utf8]{inputenc} 
\usepackage[T1]{fontenc}    
\usepackage{hyperref}       
\usepackage{url}            
\usepackage{booktabs}       
\usepackage{subfigure}
\usepackage{booktabs}
\usepackage{xcolor}
\usepackage{nicefrac}       
\usepackage{footnote}

\newcolumntype{L}[1]{>{\raggedright\let\newline\\\arraybackslash\hspace{0pt}}m{#1}}
\newcolumntype{C}[1]{>{\centering\let\newline\\\arraybackslash\hspace{0pt}}m{#1}}
\newcolumntype{R}[1]{>{\raggedleft\let\newline\\\arraybackslash\hspace{0pt}}m{#1}}

\newcommand{\ignorethis}[1]{}

\makeatletter
\DeclareRobustCommand\onedot{\futurelet\@let@token\@onedot}
\def\@onedot{\ifx\@let@token.\else.\null\fi\xspace}

\def\etc{\emph{etc}\onedot}

\makeatother

\newcommand{\myparagraph}[1]{\vspace{-8pt}\paragraph{#1}}

\titlespacing\section{0pt}{3pt plus 0pt minus 4pt}{0pt plus 0pt minus 4pt}
\titlespacing\subsection{0pt}{1pt plus 0pt minus 5pt}{0pt plus 0pt minus 6pt}

\makesavenoteenv{tabular}
\makesavenoteenv{table}
\makesavenoteenv{tabu}

\title{Learning to Design Circuits}

\author{Hanrui Wang$^*$\\
  EECS\\
  Massachusetts Institute of Technology\\
  Cambridge, MA 02139 \\
  \texttt{hanrui@mit.edu} 
  \And
  Jiacheng Yang\thanks{Equal Contribution.}\\
    EECS\\
  Massachusetts Institute of Technology\\
  Cambridge, MA 02139 \\
  \texttt{jcyoung@mit.edu} \\
  \And
  Hae-Seung Lee\\
    EECS\\
  Massachusetts Institute of Technology\\
  Cambridge, MA 02139 \\
  \texttt{hslee@mtl.mit.edu} \\
  \And
  Song Han \\
  EECS\\
  Massachusetts Institute of Technology\\
  Cambridge, MA 02139 \\
  \texttt{songhan@mit.edu}
  }
\begin{document}

\maketitle

\begin{abstract}\label{sect:abst}
    Analog IC design relies on human experts to search for parameters that satisfy circuit specifications with their experience and intuitions, which is highly labor intensive, time consuming and suboptimal. 
    Machine learning is a promising tool to automate this process. However, supervised learning is difficult for this task due to the low availability of training data: 1) Circuit simulation is slow, thus generating large-scale dataset is time-consuming; 2) Most circuit designs are propitiatory IPs within individual IC companies, making it expensive to collect large-scale datasets.
    We propose Learning to Design Circuits (L2DC) to leverage reinforcement learning that learns to efficiently generate new circuits data and to optimize circuits. We fix the schematic, and optimize the parameters of the transistors automatically by training an RL agent with no prior knowledge about optimizing circuits. After iteratively getting observations, generating a new set of transistor parameters, getting a reward, and adjusting the model, L2DC is able to optimize circuits.
    We evaluate L2DC on two transimpedance amplifiers. Trained for a day, our RL agent can achieve comparable or better performance than human experts trained for a quarter. It first learns to meet hard-constraints (eg. gain, bandwidth), and then learns to optimize good-to-have targets (eg. area, power). Compared with grid search-aided human design, L2DC can achieve $\mathbf{250}\boldsymbol{\times}$ higher sample efficiency with comparable performance. Under the same runtime constraint, the performance of L2DC is also better than Bayesian Optimization.
    
\end{abstract}
\section{Introduction}\label{sect:intro}

    Analog circuits process continuous signals, which exist in almost all electronics systems and provide important function of interfacing real-world signals with digital systems. Analog IC design has a large number of circuit parameters to tune, which is highly difficult for several reasons. First, the relationship between the parameters and performance is subtle and uncertain. Designers have few explicit patterns or deterministic rules to follow. Analog Circuits Octagon \cite{behzad2001design} characterizes strong coupled relations among performance metrics. Therefore, improving one aspect always incurs deterioration of another. A proper and intelligent trade-off between those metrics requires rich design experience and intuitions. Moreover, simulation of circuits is slow, especially for complex circuits such as ADCs, DACs, and PLLs. That makes the random search or exhaustive search impractical.
    
    There exist several methods to automate the circuit parameter search process. Particle swarm intelligence  \cite{rout2014multiobjective} is a popular approach. However, it is easy to fall into local optima in high-dimensional space and also suffers from a low convergence rate. Moreover, simulated annealing \cite{phelps2000anaconda} is also utilized,  but repeatedly annealing can be very slow and easy to fall into the local minimum. Although evolutionary algorithm \cite{liu2009memetic} can be used to solve the optimization problem, the process is stochastic and lacks reproducibility. In \cite{lyuefficient}, researchers also proposed model-based simulation-based hybrid method and utilized Bayesian Optimization \cite{pelikan1999boa} to search for parameter sets. However, the computational complexity of BO is prohibitive, making the runtime very long.
    
    Machine learning is another promising method to address the above difficulties. However, supervised learning requires large scale dataset which consumes long time to generate. Besides, most of the analog IPs are proprietary, not available to the public. Therefore, we introduce L2DC method, which leverages reinforcement learning (RL) to generate circuits data by itself and learns from the data to search for best parameters. We train our RL agents from scratch without giving it any rules about circuits. In each iteration, the agent obtains observations from the environment, produces an action (a set of parameters) to the circuit simulator environment, and then receives a reward as a function of gain, bandwidth, power, area, \etc. The observations consist of DC operating points, AC magnitude and phase responses and also transistors' states, obtained from simulation tools such as Hspice and Spectre. The reward is defined to optimize the desired Figures of Merits (FOM) composed of several performance metrics. By maximizing the reward, RL agent can optimize the circuit parameters.
    
    
    Experimental results on two different circuits environments show that L2DC can achieve similar or better performance than human experts,  Bayesian Optimization and random search. L2DC also has $\mathbf{250}\boldsymbol{\times}$ higher sample efficiency compared to grid search aided human expert design. The contributions of this paper are: 1) A reinforcement learning based analog circuit optimization method. It is a learning-based method that updates optimization strategy by itself with no need for empirical rules; 2) A sequence-to-sequence model to generate circuit parameters, which serves as the actor in the RL agent; 3) Our method achieves more than $\mathbf{250}\boldsymbol{\times}$ higher sample efficiency comparing to grid search based human design and gets comparable or better results. Under the same runtime constraint, our method can get better circuit performance than Bayesian Optimization.
    \begin{figure*}[t]
        \centering
        \includegraphics[width=0.94\textwidth]{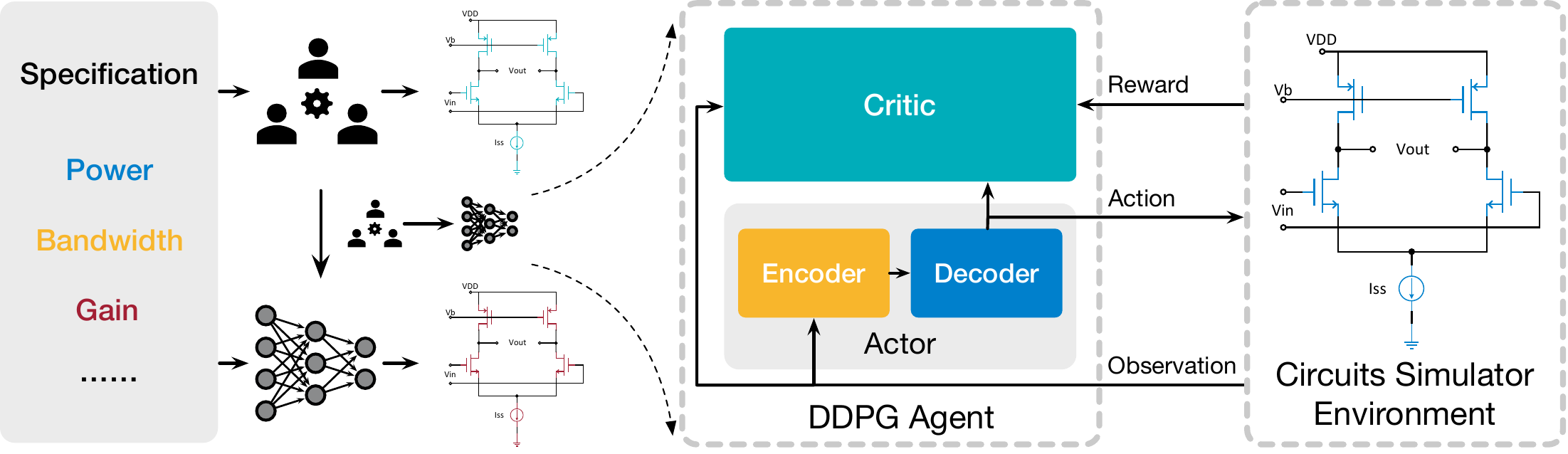}
        \caption{\label{fig:thumbnail} Learning to Design Circuits (L2DC) Method Overview.}
        \vspace{-12pt}
    \end{figure*}
\section{Methodology}
\label{sect:method}
    \subsection{Problem Definition}
    The design specification of an analog IC contain hard constraints and optimization targets. For hard constraints, designers only need to meet them with no need for any over-optimization. For optimization targets, the rule is ``the larger the better'' or ``the smaller the better''. For optimization targets, there also exist thresholds specifying the minimum performance designers need to achieve.
        
        Formally, we denote $\mathbf{x} \in \mathbb{R}^n$ as the parameters of $H$ components, $\mathbf{y} \in \mathbb{R}^m$ as the specs to be satisfied, including hard constraints and thresholds for optimization targets. We assume the mapping $\mathbf{f}: \mathbb{R}^n \rightarrow \mathbb{R}^m$ is the circuit simulator which computes $m$ performance metrics given $n$ parameters.
        We define a ratio $q$ for each spec $c$ to measure to which extent the circuit satisfies the spec. If the metric should be larger than the spec $c$, $q_c(\mathbf{x}) = f_c(\mathbf{x}) / y_c$. Otherwise $q_c(\mathbf{x}) = y_c / f_c(\mathbf{x})$. Then analog IC optimization can be formalized as a constrained optimization problem, that is, to maximize the sum of $q_c(\mathbf{x})$ of the optimization targets with all of the hard constraints being satisfied.
        
        
    \subsection{Multi-Step Simulation Environment}
        We present an overview of our L2DC method in Figure \ref{fig:thumbnail}. L2DC is able to find the optimized parameters by several epochs of interactions with the simulation environment. Each epoch contains $T$ steps. For each step $i$, the RL agent takes an observation $\mathbf{o_i}$ from the environment (it can be regarded as state $\mathbf{s_i}$ in our environments), outputs an action $\mathbf{a_i}$ and then receives an reward $r_i$. By learning from history, the RL agent is able to optimize the expected cumulative reward.
        
        The simulation of a circuit is essentially an one-step process, meaning that the state information including voltage and current of the circuit's environment cannot be directly facilitated. To effectively feed the information to RL agent, we purpose the following multi-step environment.
        
        \myparagraph{Observations}
            \begin{figure}[t]
                \centering
                \includegraphics[width=0.96\textwidth]{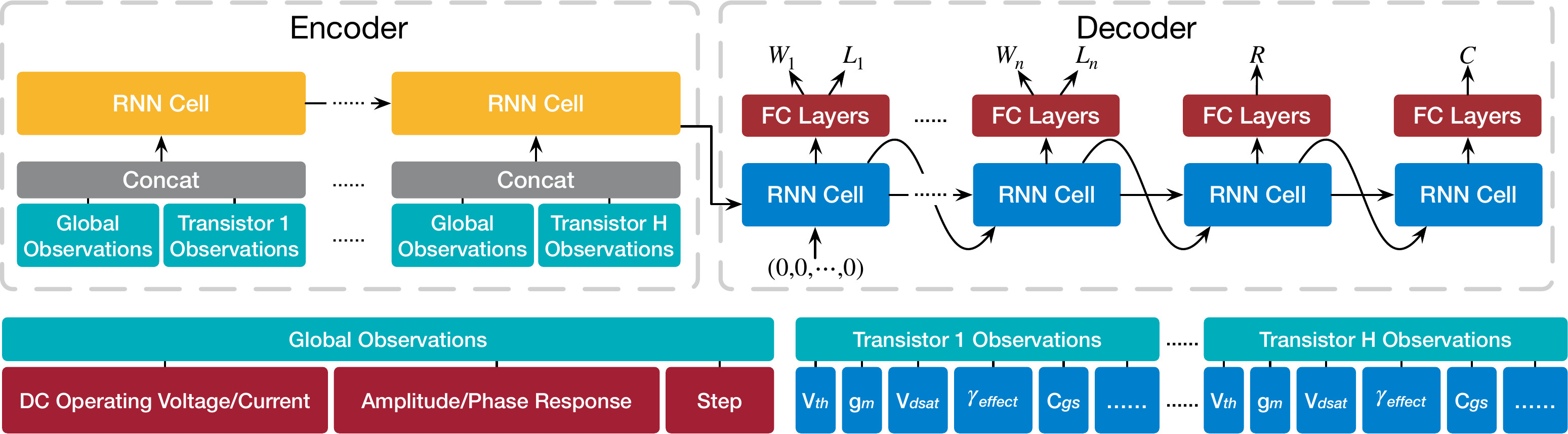}
                
                
                \vspace{-5pt}
                \caption{\label{fig:fig2} We use sequence to sequence model to generate circuit parameters (Top). Global and local observations for the RL agent (Bottom).}
                
                \vspace{-12pt}
            \end{figure}

            As illustrated in Figure \ref{fig:fig2}, at each step $i$, the observation $\mathbf{o_i}$ is separated into global observations and each transistor's own observations. The global observations 
            contain high-level features of the simulation results, including the DC operating point voltage and current at each node, AC amplitude/phase responses and a one-hot vector encoding the current environment step.
            Local observations are the features of the $i$-th transistor, containing $V_{th}, g_m, V_{dsat}, \gamma_{effect}, I_D$, capacitance between transistor pins and so on. The initial values of all global observations and local transistor observations are set to zeros.
            
        \myparagraph{Action Space}
            Supposing there are $n$ parameters to be searched, the reinforcement learning agent would output a normalized joint action $\mathbf{\overline{a_i}} \in [-1, 1]^n$ as the predicted parameters of each component at each step $i$.
            Then the actions $\mathbf{\overline{a_i}}$ are scale up to $\mathbf{a_i}$ according to the maximum/minimum constraints of each parameter $[\mathbf{pmin}, \mathbf{pmax}]$ where 
            $\mathbf{a_i}$ contains the widths and lengths of transistors, capacitance of capacitors and resistance of resistors.
            
        \myparagraph{Reward Function}
            After the reinforcement learning agent outputs circuit parameters $\mathbf{a_i}$, the simulation environment $\mathbf{f}$ will benchmark on these parameters, generating simulation results of various performance metrics. We gather the metrics together as a scalar score $d_i$. Denote $K_1(\mathbf{x})$ as the sum of $q_c$ of those hard constraints and $K_2(\mathbf{x})$ of those optimization target. Then $d_i$ is defined as
            \begin{equation}
                d(\mathbf{x}) = \begin{dcases}
                    \displaystyle K_1(\mathbf{x}) + \alpha * K_2(\mathbf{x}) + e_0 & \text{if some hard constraints are not satisfied}\\
                    \displaystyle K_2(\mathbf{x}) + e_1 & \text{if all hard constraints are satisfied}
                \end{dcases}
            \end{equation}
            When the hard-constraints in the spec are not satisfied, DDPG will optimize hard-constraint requirements and optionally optimize optimization target requirements according to the coefficient $\alpha$. When all the hard-constraints are satisfied, DDPG will only focus on optimization targets. $e_0$ and $e_1$ are two constants. They are used to make sure the scores after hard-constraints are satisfied are higher than those before hard-constraints are satisfied. To fit the reinforcement learning framework where the cumulative reward is maximized, the reward for the $i$-th step $r_i$ is defined as the finite $d_i - d_{i - 1}$.
        
    \subsection{DDPG Agent}
        As shown in Figure \ref{fig:fig2}, the DDPG \cite{Silver:2014wt, Lillicrap:2015ww} actor forms an encoder-decoder framework \cite{Sutskever:2014ty} which translates the observations to the actions. The order which we follow to feed the observations of transistors, is the order of signal propagation through a certain path from input ports to output ports, intuited by fact that the former components influence latter ones. The decoder generates transistor $W$ and $L$ in the same order as well. 
        To explore the search space, we apply truncated uniform distribution as noise to each output. Namely, $\mathbf{\tilde{a}_i} \sim \mathcal{U}(\max(\mathbf{a_i} - \boldsymbol{\sigma}, -1), \min(\mathbf{a_i} + \boldsymbol{\sigma}, 1))$, where $\boldsymbol{\sigma} \in [0, 1]$ denotes the noise volume. Besides, we also find parameter noise \cite{Plappert:2017vz} improves the performance.
        For critic network, we simply use a multi-layer perceptron estimating the expected cumulative reward of the current policy.
        

\section{Experimental Results}\label{sect:exper}

\subsection{Three-Stage transimpedance Amplifier}
\begin{figure*}[!ht]
        \centering
        \subfigure{\includegraphics[width=0.49\columnwidth, height=0.28\columnwidth]{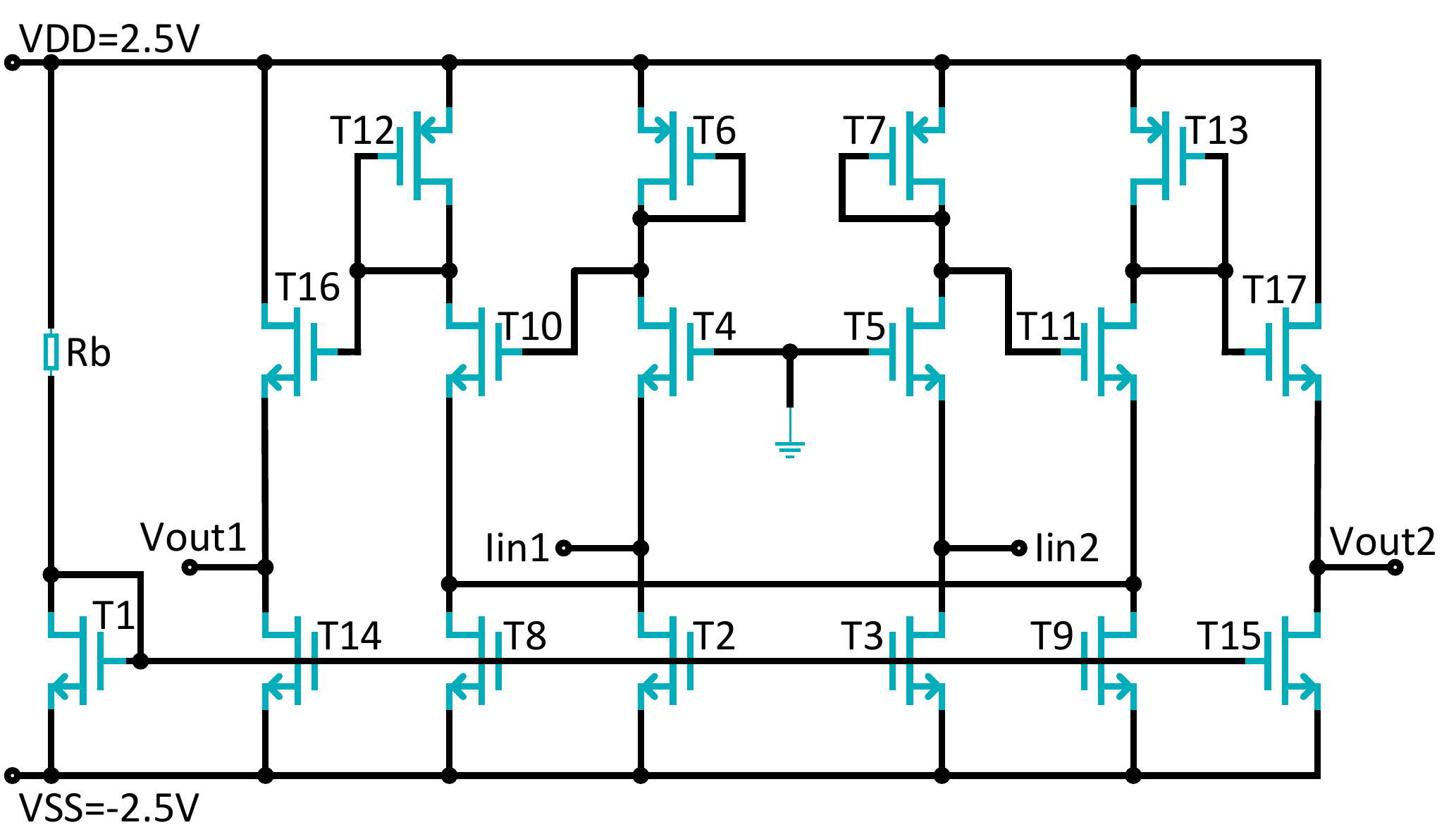}}
        \subfigure{\includegraphics[width=0.49\columnwidth, height=0.28\columnwidth]{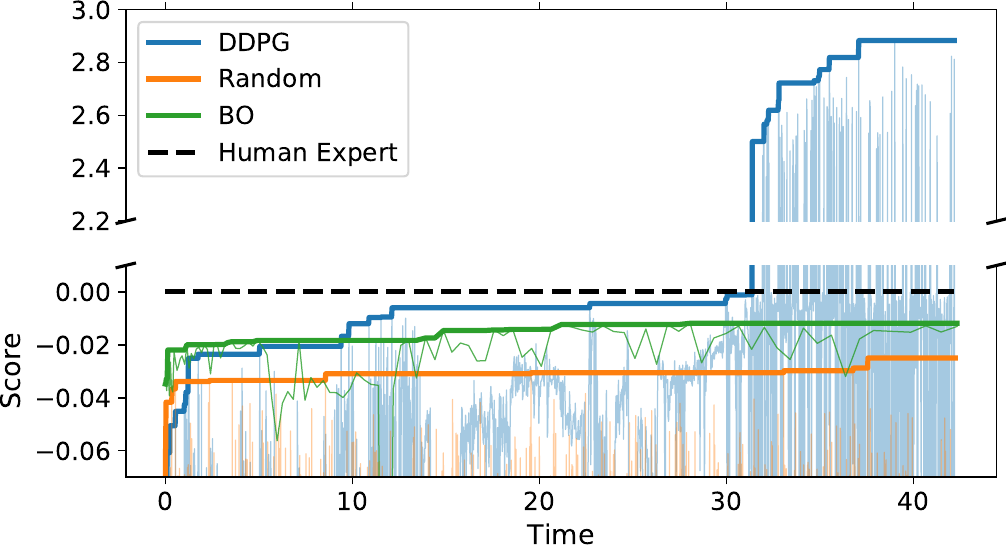}}
        \vspace{-5pt}
        \caption{\label{fig:env3}Left: Schematic of Three-stage transimpedance amplifier. 
       Right: Learning curves of three-stage transimpedance amplifier.
        }
\end{figure*}
    \begin{savenotes}
\begin{table}
    \small
    \centering
    \caption{\label{tab:env3result}Results on three-stage transimpedance amplifier. Under the same runtime constraint, random search and Bayesian Optimization cannot meet the spec hard-constraints (as marked in red); DDPG is able to satisfy all the spec hard-constraints with smallest gate area, thus achieving highest score. The sample efficiency of DDPG is 250 $\times$ higher than human design.
    }
    \begin{tabular}{c|cc|cc|cc|c}
        \toprule
            & \begin{tabular}[c]{@{}c@{}c@{}}Number of \\ Simulations \\ (Same Runtime)\end{tabular} & \begin{tabular}[c]{@{}c@{}}Sample \\ Efficiency\footnote{We ignore the sample efficiency if the spec is not met.}\end{tabular} & \begin{tabular}[c]{@{}c@{}}Bandwidth \\ ($\mathrm{MHz}$)\end{tabular} & \begin{tabular}[c]{@{}c@{}}Gain\\ ($\mathrm{k\Omega}$)\end{tabular} & \begin{tabular}[c]{@{}c@{}}Power \\($\mathrm{mW}$)\end{tabular} & \begin{tabular}[c]{@{}c@{}}Gate Area\\ ($\mathrm{\mu m}^2$)\end{tabular} & Score \\
        \hline
            \begin{tabular}[c]{@{}c@{}} Spec \cite{stanford214a} \end{tabular}  & -- & -- & 90.0 & 20.0 & 3.00 & --   & --  \\
        \hline
            Human Expert \cite{stanford214Ahuman} & 10,000,000 & 1 & 90.1 & 20.2 & \textbf{1.37} & 211 & 0.00 \\
        \hline
            Random  & 40,000 & -- & \color{red}{57.3} & 20.7 & 1.37 &  146 & -0.02 \\
        \hline
           \begin{tabular}[c]{@{}c@{}}Bayesian Opt. \cite{lyu2018batch}\end{tabular} & 1,160 \footnote{The time complexity of Bayesian Optimization is cubic to the number of samples and the space complexity is square to the number of samples. Therefore we executed BO for only 1,160 samples (the running time is the same as random and our method).} & -- &  {\color{red} 72.5} & 21.1 & {\color{red} 4.25} & 130 & -0.01 \\
        \hline
            Ours (DDPG)  & 40,000 & \textbf{250} & \textbf{92.5} & \textbf{20.7} & 2.50 & \textbf{90} & \textbf{2.88} \\
        \bottomrule
    \end{tabular}

\end{table}
\end{savenotes}
    The first environment is a three-stage transimpedance amplifier from the final project of Stanford EE214A \cite{stanford214a}. The schematic of the circuit is shown in Figure \ref{fig:env3}. We compare L2DC with random search, an grid search aided human expert design proposed by a PhD student in EE214A class as well as MACE \cite{lyu2018batch}, a Bayesian Optimization (BO) based algorithm for analog circuit optimization. The batch size of MACE is 10 and we use 50 samples for initialization.
    
    We run DDPG, BO and random search, each for about 40 hours. The comparison results are shown in Table \ref{tab:env3result}. The learning curves are shown in Figure \ref{fig:env3}. Random search is not able to meet the bandwidth hard-constraint because there are seventeen transistors making the environment very complex and design space very large. BO is also unable to meet the bandwidth and power hard-constraints. DDPG's design can meet all the hard-constraints and has slightly higher power but smaller gate area than the human expert design, so it achieves highest score. The power consumption of DDPG, though slightly higher than human design, can satisfy the course spec constraint. Moreover, the number of simulations of DDPG is $250\times$ fewer than the grid search aided human design, demonstrating high sample efficiency of our L2DC method.
    
\subsection{Two-stage transimpedance Amplifier}
    \begin{figure*}[b]
        \centering
        \subfigure{\includegraphics[width=0.49\columnwidth, height=0.28\columnwidth]{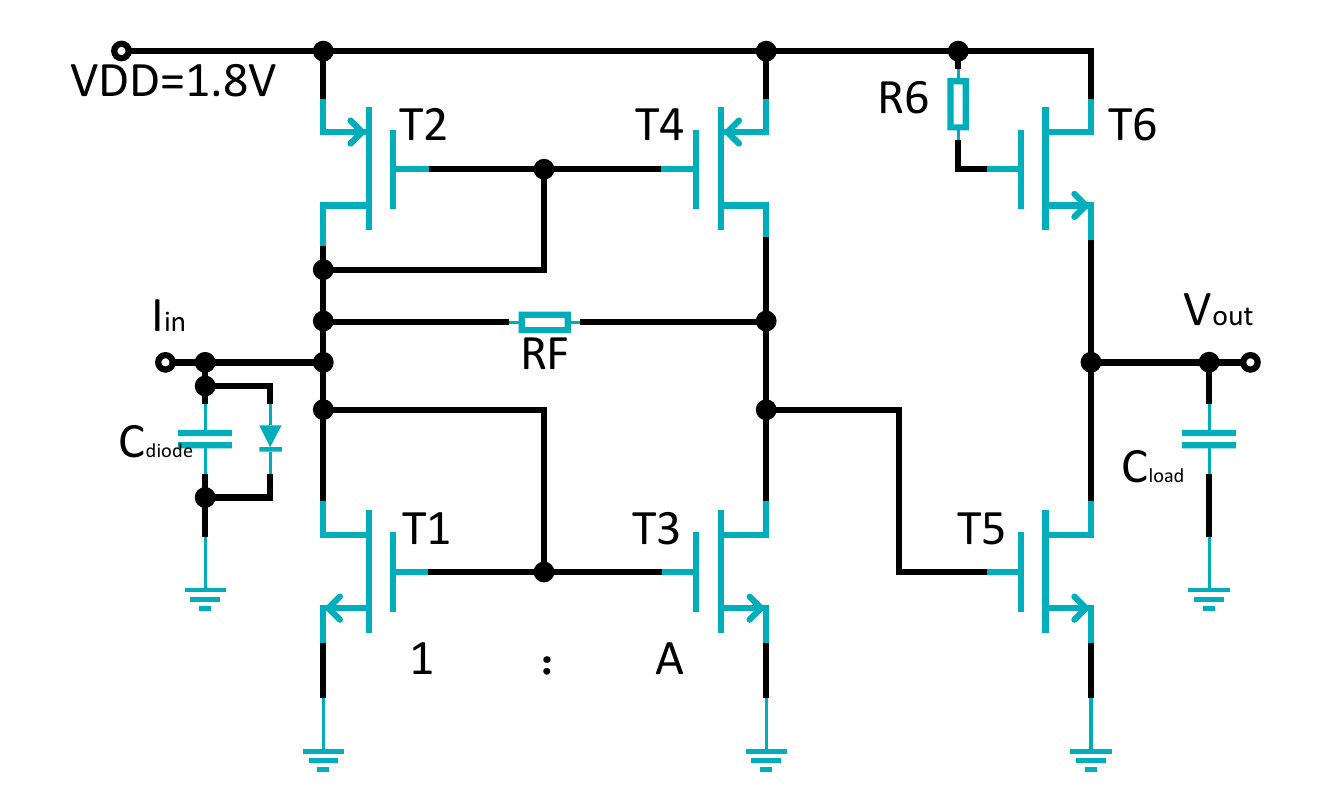}}
        \subfigure{\includegraphics[width=0.49\columnwidth, height=0.28\columnwidth]{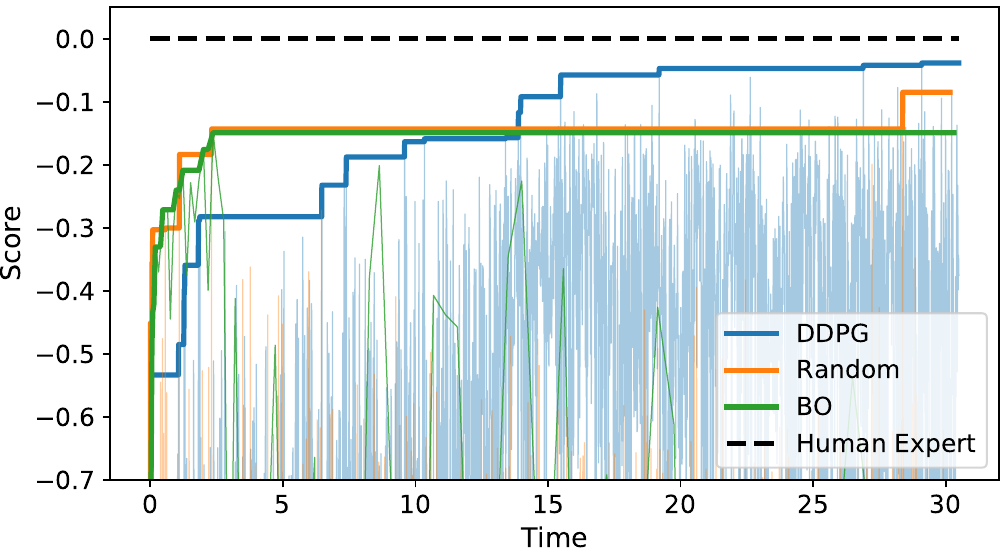}}
        \vspace{-5pt}
        \caption{\label{fig:env4}Left: Schematic of two-stage tranimpedence amplifier. Right: Learning curves of two-stage transimpedance amplifier. 
        }
    \end{figure*}
    \begin{savenotes}
\begin{table}

    \small\setlength{\tabcolsep}{3pt}
    \centering
    \caption{\label{tab:env4result}Results on two-stage transimpedance amplifier. Under the same runtime constraint, random and Bayesian Optimization are unable to meet the noise hard-constraint (as marked in red); DDPG can satisfy all the spec hard-constraints and achieve 97.143\% bandwidth of computer-aided human expert design. The sample efficiency of DDPG is 25$\times$ higher than human design.
    }
    \begin{tabular}{c|cc|ccccc|c|c}
    \toprule
       & \begin{tabular}[c]{@{}c@{}c@{}}NO. of Simu. \\(Same Runtime)\end{tabular} & \begin{tabular}[c]{@{}c@{}}Sample \\ Efficiency\end{tabular} & \begin{tabular}[c]{@{}c@{}}Noise\\ ($\mathrm{pA/\sqrt{Hz}}$)\end{tabular} & \begin{tabular}[c]{@{}c@{}}Gain\\($\mathrm{dB\Omega}$)\end{tabular} & \begin{tabular}[c]{@{}c@{}}Peaking\\(dB)\end{tabular} & \begin{tabular}[c]{@{}c@{}}Power\\($\mathrm{mW}$)\end{tabular} & \begin{tabular}[c]{@{}c@{}}Gate\\Area\\($\mathrm{\mu m}^2$)\end{tabular} & \begin{tabular}[c]{@{}c@{}}Band-\\width\\($\mathrm{GHz}$)\end{tabular} & Score \\
    \hline
        \begin{tabular}[c]{@{}c@{}} Spec \cite{stanford214Bspec} \end{tabular}  & -- & -- & 19.3 & 57.6 & 1.000 & 18.0 & -- & maximize & -- \\
    \hline
        \begin{tabular}[c]{@{}c@{}}Human\\ Expert\cite{stanford214Bhuman}\end{tabular}  & 1,289,618 & 1 & \textbf{18.6} & 57.7 & 0.927 & 8.11 & 6.17 & \textbf{5.95} & \textbf{0.00} \\
    \hline
        Random & 50,000 & -- & {\color{red} 19.8} & 58.0 & 0.488 & 4.39 & 2.93 & 5.60 & -0.08 \\
    \hline
        \begin{tabular}[c]{@{}c@{}}Bayesian\\ Opt. \cite{lyu2018batch}\end{tabular} & 880 & -- & {\color{red} 19.6} & 58.6 & 0.629 &  4.24 & 5.69 & 5.16 & -0.15 \\
    \hline
        Ours (DDPG) & 50,000 & \textbf{25} & 19.2 & \textbf{58.1} & 0.963 & \textbf{3.18} & \textbf{2.61} & 5.78 & -0.03  \\
    \bottomrule
    \end{tabular}
\end{table}
\end{savenotes}
    The second environment is a two-stage transimpedance amplifier. The schematic of the circuit is shown in Figure \ref{fig:env4}.
    The circuit is from Stanford EE214B design contest \cite{stanford214Bspec}. The contest specifies noise, gain, peaking, power as hard-constraints and bandwidth as optimization target. 
    
    We run DDPG, BO and random search, each for about 30 hours. In Table \ref{tab:env4result}, we compare DDPG result with random search, BO, and human expert design which applies a $g_m/I_D$ methodology to conduct design space search. The learning curves are shown in Figure \ref{fig:env4}. Human expert design achieves $6~\mathrm{GHz}$ bandwidth with all hard constraints being satisfied therefore receives the ``Most Innovative Design'' award. Random cannot meet the noise hard constraints. BO cannot meet the noise hard constraint either. DDPG meets all the hard constraints and achieves $5.78~\mathrm{GHz}$ bandwidth which is already $97.143\%$ of the human result, while the sample efficiency of L2DC is $25\times$ better than human expert design.
    
\section{Discussion}




\begin{figure*}[t]
    \centering
    \subfigure[Power]{\includegraphics[width=0.45\textwidth]{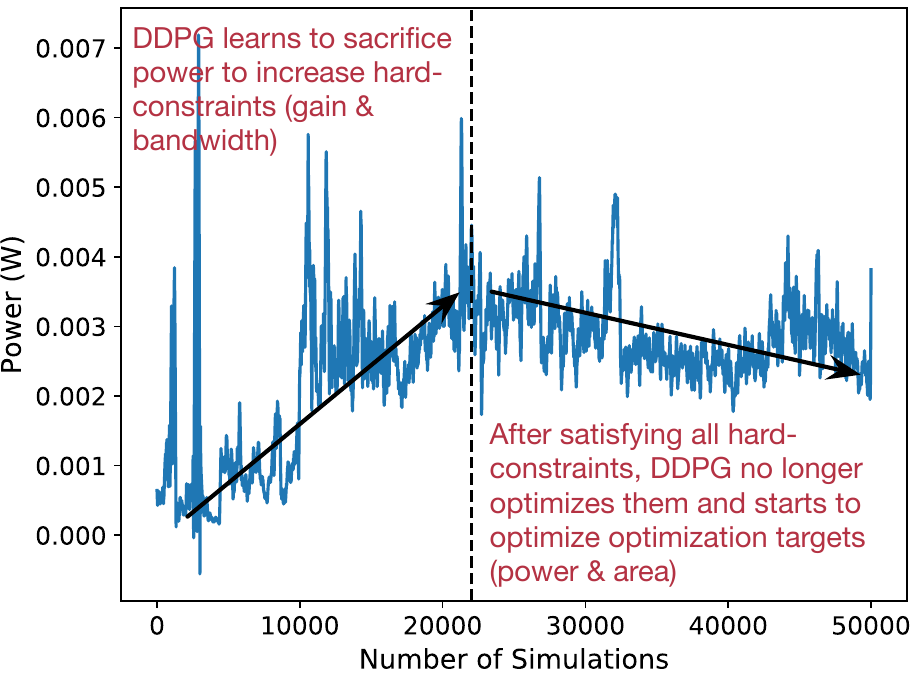}}
    \subfigure[Bandwidth]{\includegraphics[width=0.45\textwidth]{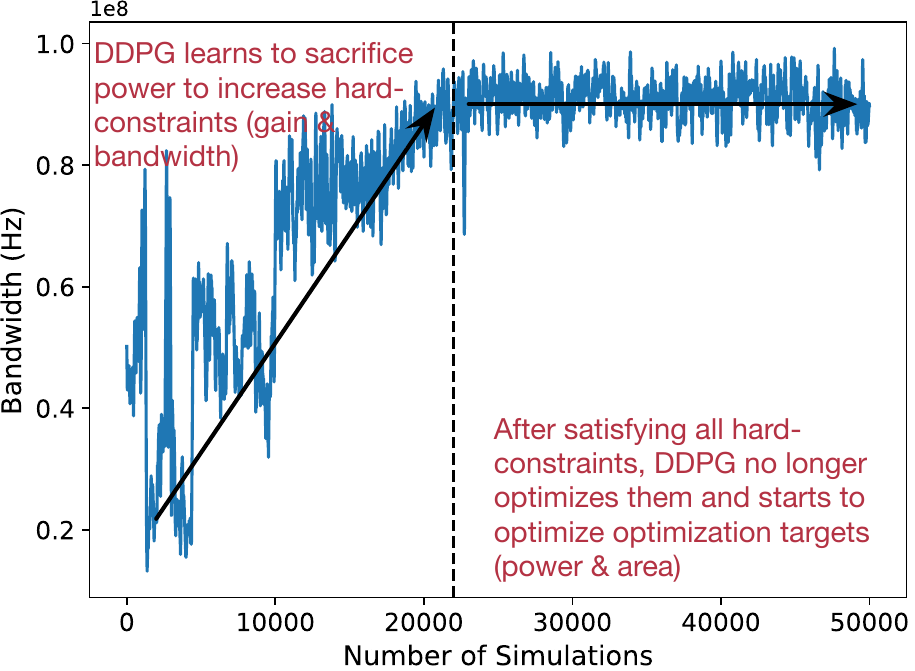}}
    \subfigure[Gain]{\includegraphics[width=0.45\textwidth]{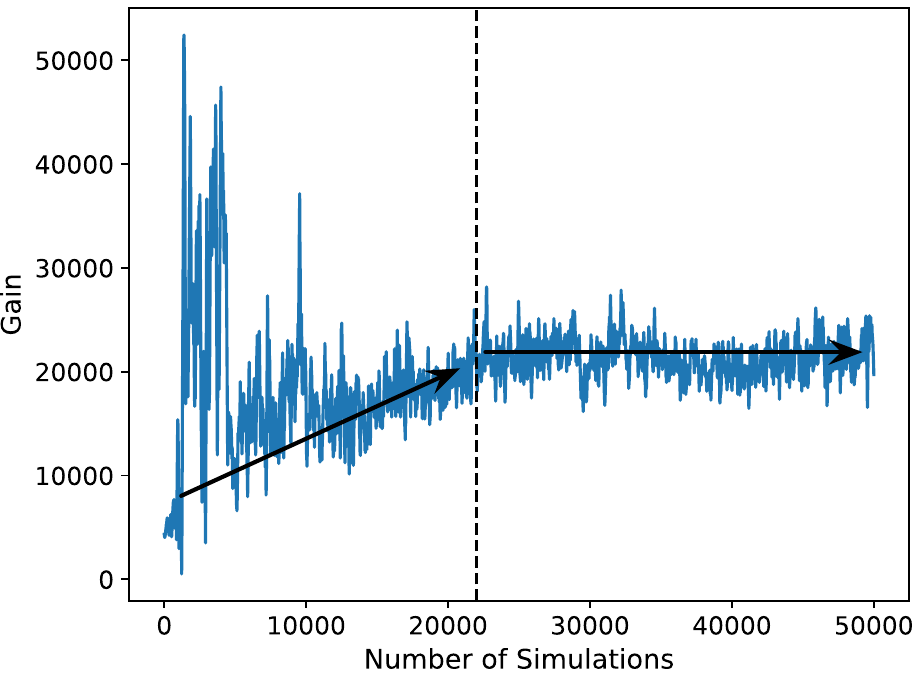}}
    \subfigure[Area]{\includegraphics[width=0.45\textwidth]{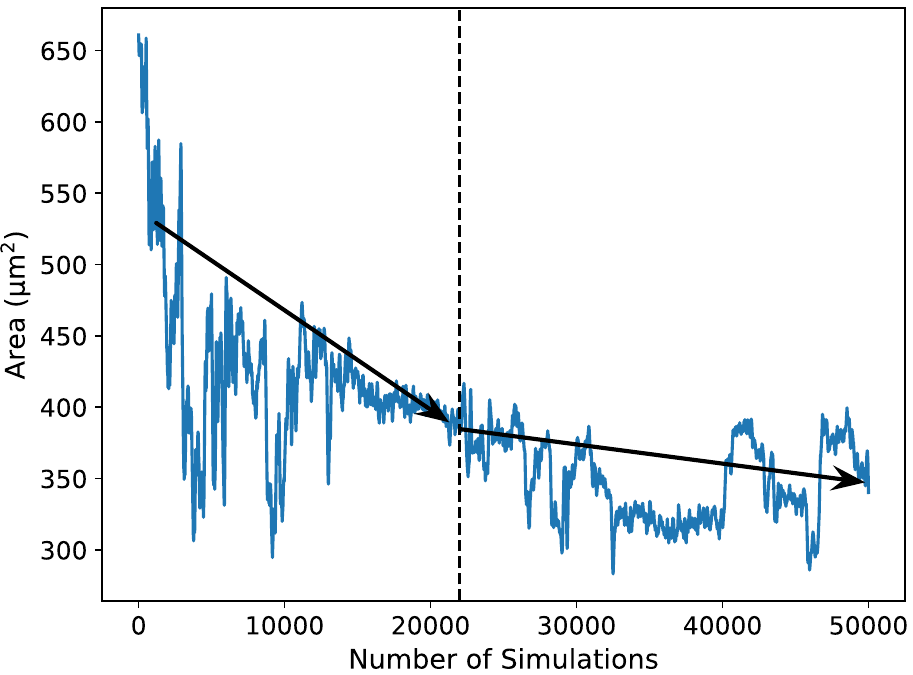}}
    
    \caption{The learning curve of the circuit RL agent. The vertical dashed line is the time when those hard-constraints (gain, bandwidth) are satisfied. RL agent learns that it should  first optimize hard-constraints (for example, obtaining more gain and more bandwidth at the cost of sacrificing more power), then improve those soft optimization targets (Fig.$(a)$ and $(d)$: decrease the power and area ) while keeping hard-constraints constant (Fig.$(b)$ and $(c)$: maintains gain and bandwidth).}
    \label{fig:learn}
\end{figure*}

As shown in Figure \ref{fig:learn}, we plot the curves of performance metrics v.s. learning steps in the three-stage transimpedance amplifier. The vertical dash line indicates the step when hard-constraints are satisfied. We can observe that power goes up and then goes down; bandwidth goes up and then stays constant; gain goes up and then remains. Therefore, from the RL agent's point of view, it firstly sacrifice power to increase hard-constraints (bandwidth and gain). After those two metrics are met, RL agents tried to keep the bandwidth and gain constant and starts to optimize power which is a soft optimization target. From this phenomenon, we can infer that RL agent has learnt some strategies in analog circuit optimization.

\section{Conclusion}
\label{sect:conc}

We propose L2DC that leverages reinforcement learning to automatically optimize circuit parameters. Comparing to supervised learning, it does not need large scale training dataset which is difficult to obtain due to long simulation time and IP issues.
We evaluate our methods on two different transimpedance amplifiers circuits. After iteratively getting observations, generating a new set of parameters by itself, getting a reward, and adjusting the model, L2DC is able to design circuits with better performance than both random search, Bayesian Optimization and human experts. L2DC works well on both two-stage transimpedance amplifier as well as complicated three-stage amplifier, demonstrating its generalization ability. Compared with grid search aided human design, L2DC can achieve comparable performance, with about $\mathbf{250}\boldsymbol{\times}$ higher sample efficiency. Under the same runtime constraint, L2DC can also get better circuit performance than Bayesian Optimization.
\section{Acknowledgements}
We sincerely thank MIT Quest for Intelligence, MIT-IBM Watson Lab for supporting our research. We thank Bill Dally for the enlightening talk at ISSCC'18. We thank Amazon for generously providing us the cloud computation resource.

\newpage
\bibliographystyle{unsrt}
\bibliography{bib/main}

\end{document}